\documentclass{article}
\usepackage{ijcai24}

\usepackage{times}
\usepackage{soul}
\usepackage{url}
\usepackage{amssymb}
\usepackage[hidelinks]{hyperref}
\usepackage[utf8]{inputenc}
\usepackage[small]{caption}
\usepackage{graphicx}
\usepackage{amsmath}
\usepackage{amsthm}
\usepackage{booktabs}
\usepackage{multirow}
\usepackage{algorithm}
\usepackage{algorithmic}
\usepackage[switch]{lineno}
\usepackage[misc]{ifsym}

\title{M2ORT: Many-To-One Regression Transformer for Spatial Transcriptomics Prediction from Histopathology Images}

\author{
Hongyi Wang$^1$
\and
Xiuju Du$^2$\and
Jing Liu$^{2}$\and
Shuyi Ouyang$^1$\and
*Yen-Wei Chen$^3$\and
*Lanfen Lin$^1$\\
\affiliations
$^1$Zhejiang University, Hanzghou, China\\
$^2$Zhejiang Lab, Hanzghou, China\\
$^3$Ritsumeikan University, Kyoto, Japan\\
\emails
chen@is.ritsumei.ac.jp
llf@zju.edu.cn,
}
\begin{document}

\maketitle

\begin{abstract}
The advancement of Spatial Transcriptomics (ST) has facilitated the spatially-aware profiling of gene expressions based on histopathology images. Although ST data offers valuable insights into the micro-environment of tumors, its acquisition cost remains expensive. Therefore, directly predicting the ST expressions from digital pathology images is desired. Current methods usually adopt existing regression backbones for this task, which ignore the inherent multi-scale hierarchical data structure of digital pathology images. To address this limit, we propose M2ORT, a many-to-one regression Transformer that can accommodate the hierarchical structure of the pathology images through a decoupled multi-scale feature extractor. Different from traditional models that are trained with one-to-one image-label pairs, M2ORT accepts multiple pathology images of different magnifications at a time to jointly predict the gene expressions at their corresponding common ST spot, aiming at learning a many-to-one relationship through training. We have tested M2ORT on three public ST datasets and the experimental results show that M2ORT can achieve state-of-the-art performance with fewer parameters and floating-point operations (FLOPs). The code is available at: https://github.com/Dootmaan/M2ORT/.
    
\end{abstract}

\section{Introduction}

\begin{figure}[t]
\includegraphics[width=0.48\textwidth]{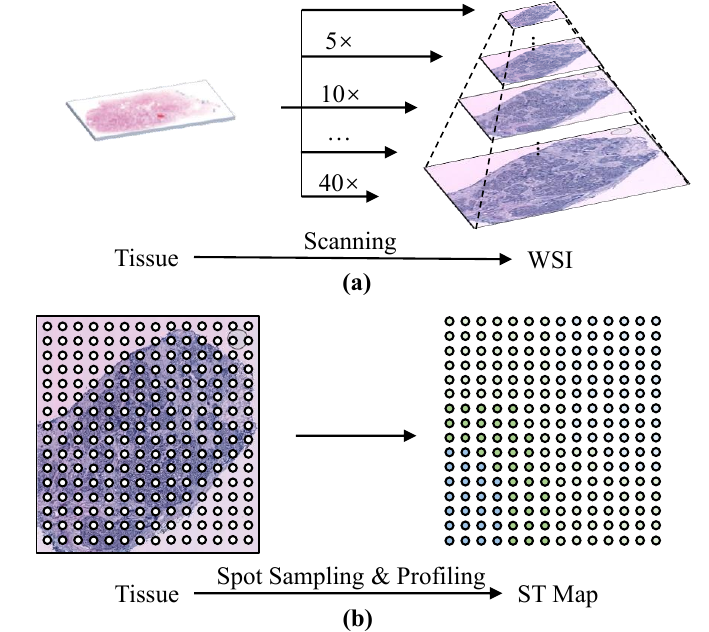}
\caption{(a) WSIs are obtained by scanning the glass slide tissues at different magnifications, resulting in a multi-scale pyramid data structure. (b) ST maps are generated by sampling spots on the glass slide tissues, followed by comprehensive profiling of gene expressions within each sampled spot. } 
% \vspace{-3mm}
\label{fig_wsi_st}
\end{figure}

Digital pathology images, as a kind of Whole Slide Images (WSIs), have witnessed widespread utilization in research nowadays as they can be more easily stored and analyzed compared to traditional glass slides \cite{niazi2019digital}. However, besides the spatial organization of cells presented in these pathology images, the spatial variance of gene expressions is also very important for unraveling the intricate transcriptional architecture of multi-cellular organisms \cite{rao2021exploring,tian2023expanding,cang2023screening}. As the extended technologies of single-cell RNA sequencing \cite{singlecellsequencing,mrabah2023toward}, ST technologies have been developed recently, facilitating spatially-aware profiling of gene expressions within tissues \cite{rodriques2019slide,lee2021xyzeq,bressan2023dawn}.

\begin{figure*}[t]
\includegraphics[width=\textwidth]{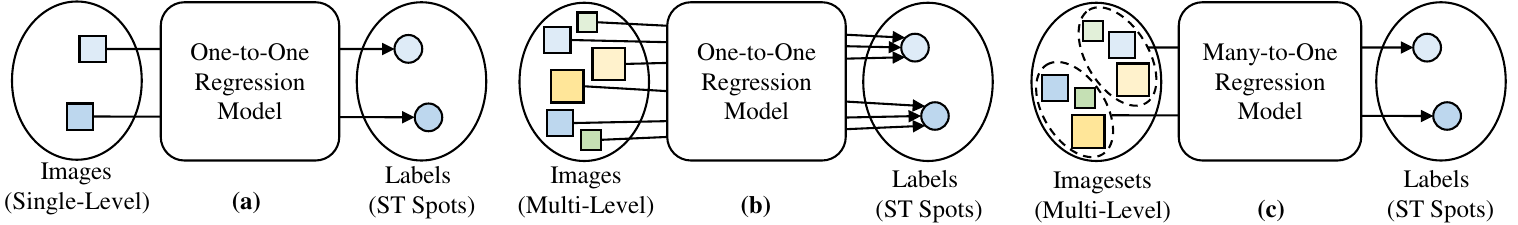}
\caption{(a) One-to-one regression models optimized with single-level image-label pairs. (b) One-to-one regression models optimized with multi-level image-label pairs. (c) Our proposed many-to-one regression model optimized with multi-level imageset-label pairs. } 
% \vspace{-3mm}
\label{m2o_vs_o2o}
\end{figure*}

A detailed illustration of the acquisition process of WSIs and ST maps is presented in Figure~\ref{fig_wsi_st}. As shown, WSIs are obtained by scanning the glass slide tissues at various magnification factors, resulting in a multi-scale hierarchical data structure \cite{ryu2023ocelot}. In a WSI, the high-resolution image scanned with maximum magnification resides at the bottom level, while the low-resolution image scanned with minimum magnification is positioned at the top level. Correspondingly, ST maps are obtained by firstly sampling spots with a fixed interval on the glass slide tissues. Each spot contains two to dozens of cells depending on different ST technologies \cite{song2021dstg}. Subsequently, the accumulated gene expressions of the cells within the spots are profiled, forming a spatial gene expression map. Despite their rapid evolution, ST technologies have yet to find widespread application in pathological analysis, primarily due to the expensive costs \cite{histogene}. In contrast, WSIs are more economical and accessible as they are routinely generated in clinics \cite{pang2021leveraging}. Consequently, there is a growing imperative to directly generate ST maps from WSIs at a low cost through deep learning methods \cite{levy2020spatial,weitz2021investigation}. 

Current approaches typically treat this ST prediction problem as a conventional regression problem \cite{stnet,deepspace}, where the network is fed with a WSI patch as input and produces the cumulative gene expression intensities of the cells within the corresponding patch area. In this paradigm, the methods are trained with single-level image-label pairs just like standard regression tasks. This makes them only able to model the relationship between the gene expressions and the images of the maximum magnification, wasting the multi-scale information inherent in WSIs (as shown in Figure~\ref{m2o_vs_o2o}(a)). Nonetheless, each level of WSIs encapsulates distinct morphological information that can be useful for ST predictions \cite{hipt,yarlagadda2023discrete}. For instance, the cell-level images can aid in evaluating the gene expressions based on the category of the cells, while images of higher levels can provide regional morphologies that can help identify the overall gene intensities through cell density information. Hence, to fully exploit the multi-scale information, we propose to address the ST prediction problem by conceptualizing it as a many-to-one modeling problem, in which case multiple images from different levels of WSI are leveraged to jointly predict the gene expressions within the spots. 

An illustrative comparison between the proposed many-to-one method and the traditional one-to-one method is depicted in Figure~\ref{m2o_vs_o2o}. As shown in Figure~\ref{m2o_vs_o2o}(b), even when images from other levels are added to the training set, the one-to-one-based model still cannot perceive the full multi-scale information at a time during training. By processing these multi-level images separately, one-to-one-based methods can only result in sub-optimal accuracy. In contrast, the many-to-one regression model is designed to comprehend the overall correlation of a collection of multi-level pathology images to their common ST spot label, thereby taking full advantage of the multi-scale information in WSIs. 
% despite that each individual sample within the collection possesses its distinct relationship to the shared label. As the value of $n$ decreases to 1, the many-to-one regression method transforms seamlessly into a one-to-one method. 
Such many-to-one-based scheme enables the joint consideration of cellular and regional morphological information, leading to a more accurate evaluation of the gene expressions. 

Finally, based on this idea, we propose M2ORT, a many-to-one-based regression Transformer designed to leverage pathology images at various levels to jointly predict the gene expressions. By incorporating the multi-scale information within the inputs, M2ORT exhibits the capability to generate more accurate ST maps. Moreover, to optimize the computational efficiency, we further introduce Intra-Level Token Mixing Module (ITMM) and Inter-Level Channel Mixing Module (ICMM) to decouple the many-to-one multi-scale feature extraction process into intra-scale representation learning and inter-scale feature fusion processes, which greatly reduces the computational cost without compromising model performance.

In summary, our contributions are: 

\begin{itemize}
    \item We propose to conceptualize the ST prediction problem as a many-to-one modeling problem, leveraging the multi-scale information embedded in the hierarchically structured WSIs for joint prediction of the ST maps.
    \item We propose M2ORT, a regression Transformer uniquely crafted to model many-to-one regression relationships for ST prediction. M2ORT enhances the precision of ST predictions with a reduced parameter count and FLOPs.
    \item In M2ORT, we propose ITMM and ICMM to decouple the multi-scale feature extraction process into intra-scale feature extraction and inter-scale feature fusion, which significantly improves the computational efficiency without compromising model performance.
    \item We have conducted thorough experiments on the proposed M2ORT method, and have proved its effectiveness with three public ST datasets.
\end{itemize}

\section{Related Works}
% \subsection{ST Prediction Based on WSIs}
The prediction of ST maps from WSIs has garnered sustained attention since the inception of ST technologies. Despite this, the literature on this specific task remains relatively scarce. ST-Net \cite{stnet} is the first work that attempts to tackle this problem. ST-Net employs a convolutional neural network (CNN) with dense residual connections \cite{resnet,densenet} to predict patch-wise gene expressions. By sequentially processing the patches in a WSI, ST-Net can eventually generate a complete ST map. Similarly, DeepSpaCE \cite{deepspace} adopts a VGG-16 \cite{vgg} based CNN for such patch-level ST prediction, and it introduces semi-supervised learning techniques to augment the training sample pool. Although these classic CNN backbones have demonstrated considerable success in various vision tasks, their performance has been eclipsed by the advancements achieved with Transformer-based models \cite{ding2023pathology}.

\begin{figure*}[t]
\includegraphics[width=\textwidth]{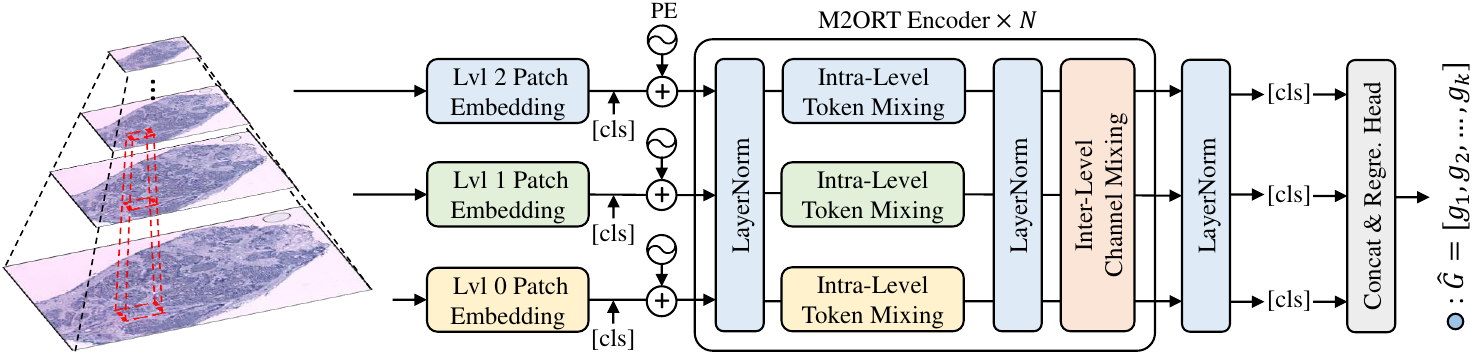}
\caption{A schematic view of the proposed M2ORT. Three patches from different WSI levels are fed into the model to jointly predict the gene expressions in the corresponding spot. PE denotes Positional Encoding in the figure.} 
% \vspace{-3mm}
\label{m2ort_structure}
\end{figure*}

As the Transformer-based methods dominate different computer vision tasks, HisToGene \cite{histogene} was proposed to leverage vision Transformers \cite{vit} for predicting ST maps. Diverging from the approach of ST-Net and DeepSpaCE, which predict one spot at a time, HisToGene proposes to predict the entire ST map at a time. HisToGene takes the sequenced patches in a WSI as network input and employs the Self-Attention mechanism \cite{transformer} to model the inter-correlations between these patches. This enables HisToGene to promptly generate predicted gene expressions for all spots simultaneously. Despite the efficiency gained from this slide-level scheme, the performance of HisToGene is constrained by the use of a relatively small ViT backbone, driven by computational limitations. Additionally, adopting patch sequences as the model input results in a reduction in the number of training samples, potentially leading to an increased risk of over-fitting.

Following the path of HisToGene, Hist2ST \cite{hist2st} was then proposed. Combining CNNs, Transformers, and Graph Neural Networks \cite{hamilton2017inductive}, Hist2ST strives to capture more intricate long-range dependencies. Like HisToGene, Hist2ST is also a slide-level method that uses the patch sequence as input to directly generate the gene expressions of all spots in an ST map. However, the complexity of its model structure results in considerable FLOPs and model size, elevating the risk of over-fitting. Contrary to the prevalent belief in the necessity of inter-spot correlations in predicting ST maps, we argue that gene expressions within a spot logically relate only to its corresponding patch area. Therefore, in our proposed M2ORT, we adhere to the patch-level scheme, predicting a single spot at a time to ensure the independence and accuracy of each prediction.

\section{Methodology}

\subsection{Problem Formulation}
Many-to-one-based regression aims at learning a mapping function from several inputs to one single output. In this work, we propose M2ORT to model such many-to-one relationships efficiently. In M2ORT, we use $I_0 \in R^{3\times H\times W}$, $I_1 \in R^{3\times \frac{H}{2}\times \frac{W}{2}}$, and $I_2 \in R^{3\times \frac{H}{4}\times \frac{W}{4}}$ to represent the three input images, where $I_i$ denotes the pathology image patch from level $i$, and $H$, $W$ represents the height and width of $I_0$ respectively. The observed gene expressions in each spot is denoted as $G=\{G^{(1)}, G^{(2)}, ..., G^{(k)}\}$, where $k$ is the number of genes. The goal is to minimize the mean squared error (MSE) between $\hat{G}=\mathrm{M2ORT}(\{I_0,I_1,I_2\}|\theta)$ and $G$ by optimizing the network parameters $\theta$.

\begin{figure}[t]
\includegraphics[width=0.48\textwidth]{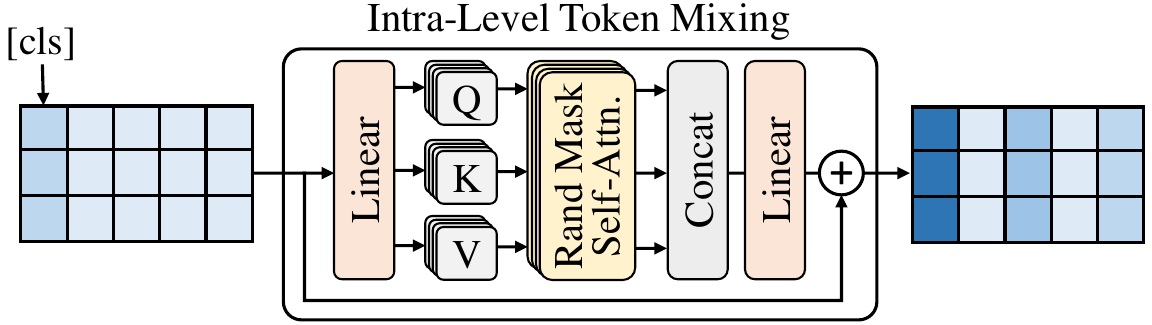}
\caption{The network structure of ITMM. This module needs to be applied to each level's sequence separately.} 
% \vspace{-3mm}
\label{Intra_level_mixing}
\end{figure}

\begin{figure}[t]
\includegraphics[width=0.48\textwidth]{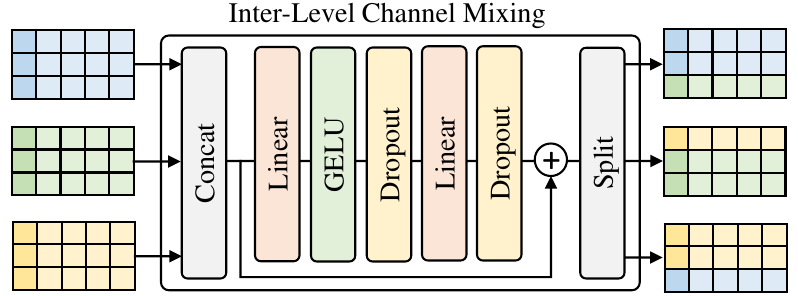}
\caption{The network structure of ICMM.} 
% \vspace{-3mm}
\label{inter_level_mixing}
\end{figure}

\subsection{Overview of M2ORT}
A schematic view of the proposed M2ORT is presented in Figure~\ref{m2ort_structure}. Upon receiving the multi-scale pathology image patches from three different levels, M2ORT initially sends them into their respective patch embedding layers, where the lower level image uses a larger patch size and the higher level image uses a smaller patch size. In this way, their sequence length can be identical, which can facilitate more effective cross-level information exchange during training. After appending $[cls]$ token to each sequence, a global-wise Layer Normalization \cite{layernorm} is performed on the three sequences to maintain consistent numerical ranges in their representations. Then, to decouple the multi-scale feature extraction process, we first perform intra-scale representation learning within each sequence using ITMM. After that, ICMM is introduced to facilitate cross-scale information exchange between different levels of inputs. This multi-scale feature extraction module is termed the M2ORT Encoder and is iterated $N$ times within M2ORT. Finally, the three $[cls]$ tokens are concatenated to be fed into the regression head for the ST spot prediction.

\subsection{Level-Dependent Patch Embedding}
To facilitate seamless cross-scale interaction within M2ORT, the embedded sequences from different levels's input images should be identical. Hence, we introduce Level-Dependent Patch Embedding (LDPE) to realize this goal. LDPE uses patch size $p$ for $I_0$, $\frac{p}{2}$ for $I_1$, and $\frac{p}{4}$ for $I_2$ to convert the three input images into unified $L\times C$ sequences $S_0,S_1,S_2$, where $L=\frac{HW}{p^2}$ represents the sequence length and $C$ denotes the number of channels after embedding. Each level's patch embedding module is composed of two linear layers and a GELU activation function to non-linearly project the flattened patches into $C$-dimension feature vectors.

\subsection{Intra-Level Token Mixing}
After patch embedding, a $[cls]$ token is appended to the beginning of each sequence, changing the sequence length from $L$ to $L+1$. Then, a fully-learnable positional encoding is added to each sequence to encode the positional information. After that, each level's sequence data undergoes processing through ITMM to extract the intra-level features, whose structure is depicted in Figure~\ref{Intra_level_mixing}. Based on the sequence data, $Q$, $K$, and $V$ matrices are projected by a linear layer for the subsequent Multi-Head Self-Attention. It should be noted that M2ORT introduces Random Mask Self-Attention (RMSA) to enhance its generalization ability, wherein some self-attention scores are randomly masked by 0 with a probability of $m$. RMSA can be mathematically defined as:
\begin{equation}
    RMSA(Q,K,V,m)=M(m)\odot Softmax(\frac{QK^T}{\sqrt{d_K}})V,
\end{equation}
\noindent where $M(m)\in R^{(L+1)\times (L+1)}$ is a binary mask, with each element having a probability of $m$ to be 0 and $1-m$ to be 1. The $\odot$ denotes the element-wise multiplication.

After concatenating the outputs from different heads, an additional linear layer is appended and the residual connection is added before the output can be sent into the subsequent modules. 

\subsection{Inter-Level Channel Mixing}

After the inter-level feature maps have been properly exploited, these acquired representations are amalgamated into ICMM for cross-level information exchange. Given that the sequence length of $S_0$, $S_1$, and $S_2$ are identical, these data can be effortlessly concatenated along the channel dimension to form a $(L+1)\times 3C$ multi-scale representation map. Based on the concatenated sequence, channel-wise mixing is executed by two linear layers and a GELU activation layer. The two linear layers establish an information bottleneck, where the first linear layer projects the $3C$ channel to $2C$ latent space and the next linear layer restores the information from $2C$ to $3C$. This strategy aids in mitigating redundant cross-level features within different levels' sequences, enhancing the efficiency of cross-level feature fusion. Additionally, Dropout \cite{dropout} is also used in this module to increase model generalization ability. However, since the subsequent M2ORT Encoder still requires multiple independent sequences as input, the concatenated feature maps still need to be split into three $(L+1) \times C$ sequences along the channel dimension.

\subsection{Regression Head}
Upon the completion of $N\times$ M2ORT Encoders, the multi-scale features are properly handled. Next, the regression head is applied on the three $[cls]$ tokens to predict the gene expression of their corresponding common ST spot. The current M2ORT model uses one linear layer on the fused $[cls]$ token to project the $1 \times 3C$ feature vector into $k$-channel gene expression vector $\hat{G}$. MSE loss is then computed and back-propagated to optimize the network parameters $\theta$ in an end-to-end manner. 
 
\section{Experiments}

\subsection{Datasets and Metrics}
\subsubsection{Datasets}
In our experiments, we utilize three public datasets to evaluate the performance of the proposed M2ORT model.

The first dataset is the human breast cancer (HBC) dataset \cite{dataset1}. This dataset contains 30,612 spots in 68 WSIs, and each spot has up to 26,949 distinct genes. The spots in this dataset exhibit a diameter of 100 µm, arranged in a grid with a center-to-center distance of 200 µm.

The second dataset is the human HER2-positive breast tumor
dataset \cite{dataset2}. This dataset consists of 36 pathology images and 13,594 spots, and each spot contains 15,045 recorded gene expressions. Similar to the previous dataset, the ST data in this dataset also features a 200µm center-to-center distance between each captured spot with the diameter of each spot also being 100µm. 

The third dataset is the human cutaneous squamous cell carcinoma (cSCC) dataset \cite{dataset3}, which includes 12 WSIs and 8,671 spots. Each spots in this dataset have 16,959 genes profiled. All the spots have a diameter of 110µm and are arranged in a centered rectangular lattice pattern with a center-to-center distance of 150µm. 

\subsubsection{Metrics}
We employ the mean values of Pearson Correlation Coefficients (PCC) and Root Mean Squared Error (RMSE) of the spots to evaluate the regression accuracy. Mathematically, PCC can be described as:
\begin{equation}
PCC=\frac{Cov (G, \hat{G})}{Var (G) \cdot Var (\hat{G})},
\end{equation}

\noindent where $Cov(\cdot)$ is the covariance, and $Var(\cdot)$ is the variance.

Subsequently, RMSE can be mathematically defined as:
\begin{equation}
RMSE=\sqrt{\frac{1}{k}\sum^k_{i=1}(G^{(i)}-\hat{G}^{(i)})^2},
\end{equation}

\noindent where $G$ is the ground truth gene expressions of a spot, $\hat{G}$ is the corresponding predicted result, $k$ is the total number of the genes in $\hat{G}$, and $\hat{G}^{(i)}$ denotes the $i$-th gene in $\hat{G}$.

\subsection{Implementation Details}

Using the same pre-processing procedures in \cite{hist2st}, we normalize the gene expression counts for each spot by dividing them by their total counts and multiplying the result by the scale factor 1,000,000. Then, the values are natural-log transformed by log(1 + $x$) where $x$ is the normalized count. Given the high sparsity of ST maps, we select 250 spatially variable genes for regression on each dataset. 

For all datasets, we use a patch size of 224$\times$224 for each matching spot on level 0 pathology image, and the patch size $p$ for level 0 patch embedding is set to 16 accordingly. In each dataset, 60\% of the WSIs and their corresponding ST maps are used for training, 10\% for validation, and the remaining 30\% for testing. All the methods are trained with Adam \cite{adam} optimizer with a learning rate of 1e-4 for 100 epochs. Batch size is 96 for patch-level methods and 1 for slide-level methods. 
% All the methods are trained on two Nvidia Tesla M40 GPUs (12GB$\times$2).

With varying network configurations, we introduce three variants of M2ORT for our experiments, namely M2ORT-\underline{S}mall, M2ORT-\underline{B}ase, and M2ORT-\underline{L}arge. The detailed configurations of them are presented in Table~\ref{m2ort_variants}. We recommend using M2ORT-B under most circumstances as it strikes a balance between model performance and computational cost. M2ORT-S has a lightweight design, and thus is recommended for devices with very limited computational resources. Conversely, M2ORT-L, with a comparatively larger model size, is recommended when there is ample training data.

\begin{table}[t]
\center
\begin{tabular}{@{}ccccc@{}}
\toprule
Method      & N  & n\_head & C   & Param\# (M) \\ \midrule
M2ORT-\underline{S}mall & 6  & 3        & 192 & 6.83         \\
M2ORT-\underline{B}ase  & 8  & 4        & 256 & 15.38    \\
M2ORT-\underline{L}arge & 12 & 6        & 384 & 49.89  \\ \bottomrule
\end{tabular}
\caption{Variants of M2ORT by adjusting the repeated times $N$ of M2ORT Encoder, number of head $n\_head$ in ITMM, and the channels $C$ for LDPE.}
\label{m2ort_variants}
\end{table}

\begin{table}[t]
\renewcommand\tabcolsep{2pt}
\begin{tabular}{@{}ccccccccc@{}}
\toprule
\multicolumn{3}{c}{Input} & \multicolumn{2}{c}{HBC} & \multicolumn{2}{c}{HER2+} & \multicolumn{2}{c}{cSCC} \\ \cmidrule(l){1-3}\cmidrule(l){4-5} \cmidrule(l){6-7} \cmidrule(l){8-9} 
Lvl 0   & Lvl 1  & Lvl 2  & PCC        & RMSE       & PCC         & RMSE        & PCC        & RMSE        \\ \midrule
\checkmark  & & & 47.10        &  3.15       &   43.33            &  3.05         &   49.34           &   3.60        \\
 & \checkmark &        &   45.73          &  3.17          &   43.12          & 3.05            &  48.78          &  3.81  \\
 &        & \checkmark &  41.62          &  3.06          &   40.24          & 3.07            &  45.78          &  3.89           \\ \midrule
 \checkmark & \checkmark & &   47.98   &  3.15    &   43.85        &  2.99      &   50.33   &   3.55   \\
\checkmark &  & \checkmark   &    47.44   &  3.17     &   43.61       &  2.99  &   49.73   &   3.61   \\
 & \checkmark & \checkmark &    46.07     &  3.14  &   43.45          & 3.04   &  49.11  &  3.80    \\ \midrule
\checkmark & \checkmark & \checkmark & 48.59        & 3.13         &  44.37           & 2.89          &  50.97  &  3.37  \\
        \bottomrule
\end{tabular}
\caption{Abaltion study results on the inputs of M2ORT. PCC is reported in percentage numbers.}
\label{m2o_ablation}
\end{table}

\subsection{Ablation Study}
\subsubsection{Study on the Many-To-One and One-To-One Schemes}
\label{sec:m2o_v_o2v}
To assess the effectiveness of the many-to-one-based modeling scheme, we have conducted thorough ablation studies. Using M2ORT-B as the backbone, various input combinations were fed into the model for performance evaluation. The experimental results are summarized in Table~\ref{m2o_ablation}. 
% It is worth noting that, despite variations in the number of inputs, the model width and depth remained constant under all circumstances. This ensures a fair comparison of experimental results.

Analysis of the table reveals that when utilizing only one single input image, i.e., employing M2ORT as an one-to-one-based method, using level 0 pathology images yields optimal results across all three datasets. This superiority is attributed to the comprehensive high-frequency information present in the level 0 pathology images, which is absent in higher-level pathology images. However, it is also shown that relying solely on level 0 images as input results in inferior performance compared to many-to-one-based schemes. Specifically, simultaneous utilization of level 0 and level 1 images during M2ORT training yields performance improvements of 0.88\%, 0.52\%, and 0.99\% in PCC on HBC, HER2+, and cSCC datasets, respectively. After introducing level 2 images as the third input, the performance of M2ORT further increases to 48.59\%, 44.37\%, and 50.97\% PCC on the three datasets, which substantiates the effectiveness of the proposed many-to-one-based modeling scheme. 

\begin{table}[t]
\renewcommand\tabcolsep{1.5pt}
\center
\begin{tabular}{@{}cccccccc@{}}
\toprule
\multirow{2}{*}{\begin{tabular}[c]{@{}c@{}}M2ORT\\Variant\end{tabular}} & \multirow{2}{*}{ITMM} & \multirow{2}{*}{ICMM} &\multicolumn{3}{c}{PCC (\%)} & \multirow{2}{*}{\begin{tabular}[c]{@{}c@{}}Param\#\\(M)\end{tabular}} & \multirow{2}{*}{\begin{tabular}[c]{@{}c@{}}FLOPs\\(G)\end{tabular}} \\ \cmidrule(lr){4-6} 
                         &         &                   & HBC       & HER2+        &   cSCC                &                             \\ \midrule
\multirow{3}{*}{Small} & \checkmark  & \checkmark & 47.78  & 43.98          & 50.70       & 6.83     & 1.44                       \\
 & $\times$  & \checkmark  & 47.29    &   44.13        &  50.77          & 11.47     & 2.23                       \\ 
  & $\times$  & $\times$  & 47.03    &   43.77        &  50.39           & 30.07     & 5.89                       \\
  \midrule
\multirow{3}{*}{Base} & \checkmark   &  \checkmark   & 48.59       &   44.37   &  50.97       & 15.38             & 3.30                       \\
   &    $\times$  & \checkmark  & 48.62     &  44.28 &  50.81           & 26.39     & 5.16                       \\   &    $\times$  & $\times$  & 48.46     &  44.03 &  50.42           & 57.87     & 11.35                       \\ \midrule
\multirow{3}{*}{Large} & \checkmark   &  \checkmark      & 48.90   & 44.61 & 51.21       & 49.89       & 10.81    \\
  &     $\times$  & \checkmark   & 48.93   &  44.67 & 51.13   & 87.05   & 17.08                      \\
   &     $\times$  & $\times$  & 48.71   &  44.50 & 50.88   & 150.78   & 29.63                      \\
  \bottomrule
\end{tabular}
\caption{Ablation study results about the design of M2ORT Encoder, the multi-scale feature extractor in M2ORT.}
\label{ablation_decouple}
\end{table}

\begin{table}[t]
\begin{tabular}{@{}ccccccc@{}}
\toprule
\multirow{2}{*}{\begin{tabular}[c]{@{}c@{}}Value \\of $m$\end{tabular}} & \multicolumn{2}{c}{HBC} & \multicolumn{2}{c}{HER2+} & \multicolumn{2}{c}{cSCC} \\ \cmidrule(l){2-3} \cmidrule(l){4-5} \cmidrule(l){6-7}
   & PCC        & RMSE       & PCC         & RMSE        & PCC        & RMSE        \\ \midrule
0       & 48.48        & 3.14         &  44.31           & 2.97          &  50.94  &  3.39      \\
0.1          & 48.59        & 3.13         &  44.37           & 2.89          &  50.97  &  3.37    \\
0.2       &   48.57         &  3.13          &  44.37           & 2.89            &    50.99 & 3.37            \\
0.3      &   48.53         &  3.14   &  44.33     &  2.91        &    51.02        &  3.36           \\
0.4    &  48.50   &   3.14         &   44.31   &   2.91          &   51.17   & 3.36            \\
0.5     & 48.39        & 3.16         &  44.29           & 2.93          &  51.11  &  3.38     \\ \bottomrule
\end{tabular}
\caption{Parameter study results of $m$ on M2ORT-B. PCC is reported in percentage numbers. }
\label{parameter_study}
\end{table}

\begin{table*}[t]
\centering
\begin{tabular}{@{}ccccccccc@{}}
\toprule
\multirow{2}{*}{Methods} & \multicolumn{2}{c}{HBC} & \multicolumn{2}{c}{HER2+} & \multicolumn{2}{c}{cSCC} & \multirow{2}{*}{\begin{tabular}[c]{@{}c@{}}Parameter\\ Count (M)\end{tabular}} & \multirow{2}{*}{\begin{tabular}[c]{@{}c@{}}FLOPs \\(G)\end{tabular}} \\ \cmidrule(lr){2-7}
                         & PCC(\%)      & RMSE     & PCC(\%)       & RMSE      & PCC(\%)      & RMSE      &                                                                                &                                                                                  \\ \midrule

ResNet50 \cite{resnet}                & 47.10        &  3.17        &   43.33            &  3.04         &   49.34           &   3.60        & 24.02    &  4.11                                                                                \\
ViT-16/B \cite{vit}  & 46.67  &  3.17  & 43.78   &  3.09  &  49.01   & 3.77  & 57.45  &  11.27  \\ \midrule
DeepSpaCE \cite{deepspace}                 & 46.01        & 3.19         &  42.57             &   3.17        & 48.99             & 3.73          & 135.29   & 15.48   \\
ST-Net \cite{stnet}    &  47.78 & 3.16         & 43.01  & 3.07          &  49.37   &  3.58 & \underline{7.21}    &   \underline{2.87}   \\
HisToGene \cite{histogene}               &   44.76 & 3.20 & 36.97         &   3.62         &   45.71 &  3.93         &    187.99   &  135.07  \\
Hist2ST \cite{hist2st}  &    45.00  &   3.18  &  40.02  & 3.06      &  46.71   & 3.88     &  675.50  &  1063.23 \\ \midrule
M2ORT-\underline{S}mall (Ours)            & 47.78      & 3.13        &   43.98            & 2.89          & 50.70       &  3.38       & \textbf{6.83}        & \textbf{1.44}                                                                            \\
M2ORT-\underline{B}ase (Ours)            & \underline{48.59}        & \underline{3.13}         &  \underline{44.37}           & \underline{2.89}          &  \underline{50.97}  &  \underline{3.37}         & 15.38    & 3.30   \\ 
M2ORT-\underline{L}arge (Ours)            &   \textbf{48.90}     &   \textbf{3.10}       &   \textbf{44.61}            &  \textbf{2.87}         &  \textbf{51.21}            &  \textbf{3.34}        & 49.89   &  10.81                                                                          \\ \bottomrule
\end{tabular}
\caption{Experimental results of comparing M2ORT series with other methods. The best results are marked in bold, and the second-best results are underlined. }
\label{comparison_with_other_methods} 
\end{table*}

\subsubsection{Study on the Decoupling Mechanism in M2ORT Encoder}

M2ORT Encoder decouples the multi-scale feature extraction process into intra-scale feature extraction and inter-scale feature fusion through ITMM and ICMM respectively. To verify the effectiveness and efficiency of the decoupling design, we have conducted an ablation study on the M2ORT Encoder. Experimental results of this study are presented in Table~\ref{ablation_decouple}.

% As shown, the original M2ORT Encoder achieves 47.78\%, 48.59\%, and 48.90\% PCC on HBC dataset with the three M2ORT variants. 
First, we change ITMM into directly extracting features from the concatenated input sequences, destroying the decoupling design. In this case, the parameter count and FLOPs nearly doubled while the performance generally remains unchanged on all variants and all datasets. Then, we remove ICMM and cancel the bottleneck design, changing to use the same feed-forward block in ViT-B/16. This time, the parameter count and FLOPs doubled again, and the performance even dropped a little due to the harmed generalization ability. In summary, ITMM and ICMM improve the computational efficiency by over 4$\times$, proving the effectiveness of them for decoupling the multi-scale feature extraction process.

\subsubsection{Parameter Study on RMSA}

RMSA randomly masks self-attention scores based on the hyper-parameter $m$. A larger value of $m$ might impede the learning of long-range dependencies, while a smaller $m$ may result in RMSA degrading to normal self-attention. 
Hence, to identify the optimal value, A parameter study is conducted based on M2ORT-B, and the results are presented in Table~\ref{parameter_study}.

The findings from the table indicate that the random masking scheme enhances regression accuracy across all three datasets. Notably, the model's performance exhibits robustness to variations in the parameter $m$. Even when masking 50\% of the self-attention scores at random, M2ORT-B demonstrates exceptional learning of the many-to-one regression relationships, showcasing its remarkable stability. We have also observed that the optimal value of $m$ may vary a little across different datasets due to the different training set scales. Generally, we recommend using $m$=0.1 in most cases, and all experimental results of M2ORT series in this paper are reported under the setting of $m$=0.1.

% However, we have also observed that the optimal value of $m$ may vary across different datasets. For instance, in the HBC dataset and HER2+ dataset, the model achieves the best prediction accuracy when $m$=0.1. Conversely, for the smaller cSCC dataset, the model's performance peaks at $m$=0.4. This discrepancy is primarily attributed to varying training set scales. Larger-scale datasets, such as HBC, tend to exhibit lower over-fitting probabilities, thus requiring a lower masking rate to enhance generalization ability. On the other hand, smaller-scale datasets, like cSCC, may necessitate introducing more disturbance during training to mitigate over-fitting risks. Generally, we recommend using $m$=0.1 in most cases, and all experimental results for the M2ORT series in this paper are reported under the setting of $m$=0.1.

\begin{figure}[t]
\includegraphics[width=0.48\textwidth]{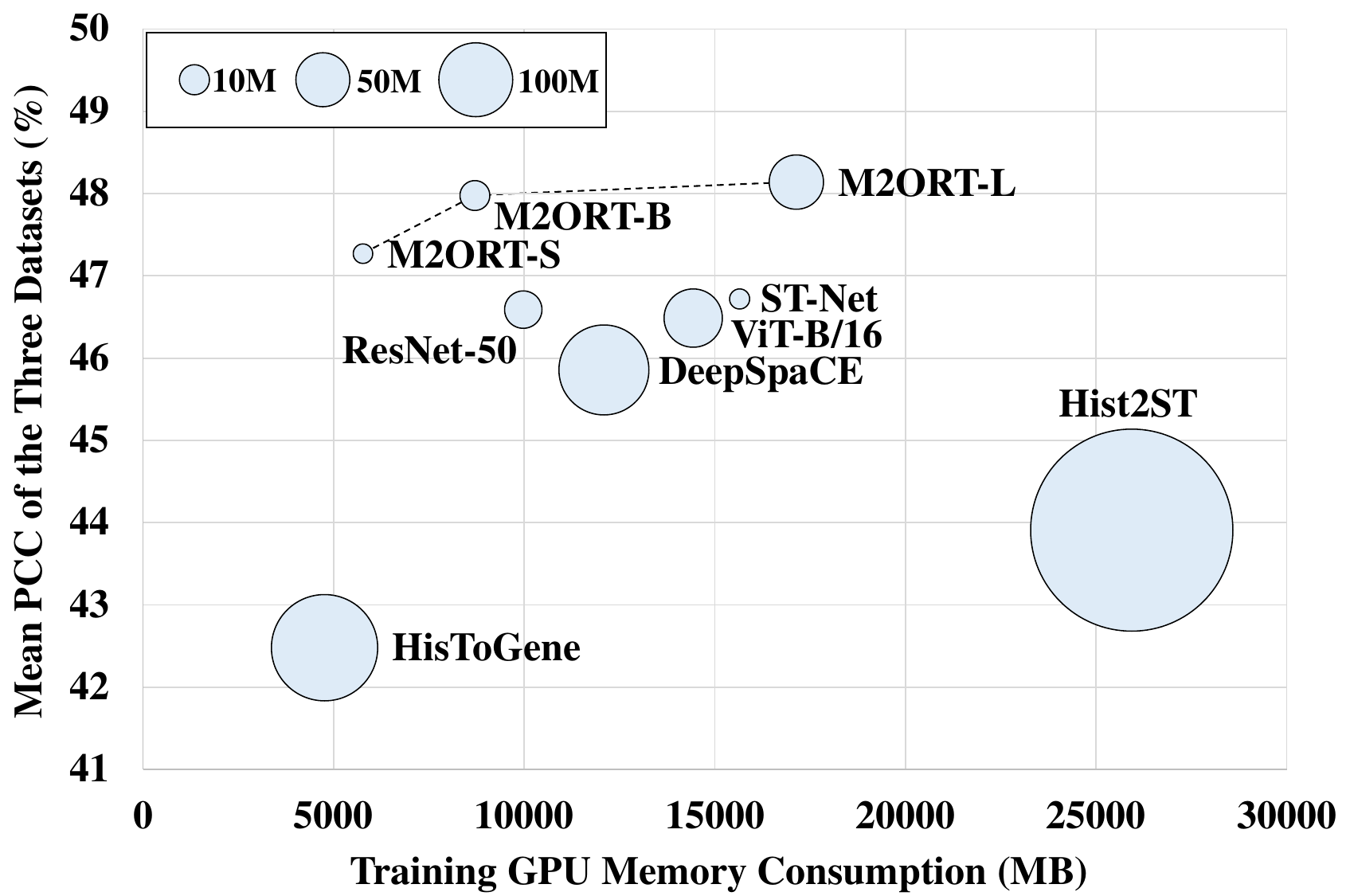}
\caption{Visual comparison of different methods. Methods in the upper area have a higher mean PCC over the three datasets, while those in the more left area consume less GPU memory during training. The area of the circle indicates the parameter count of the corresponding method.} 
% \vspace{-3mm}
\label{performance_visual_comparison}
\end{figure}

\begin{figure*}[t]
\includegraphics[width=\textwidth]{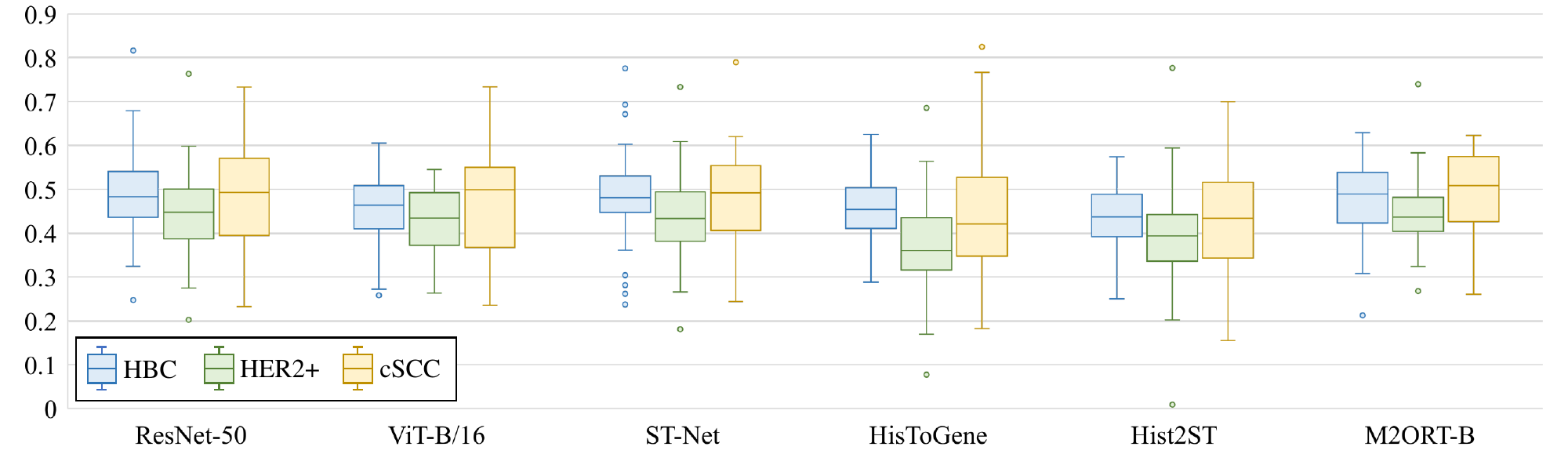}
\caption{The box-plot of different methods' test PCC on the three datasets.} 
% \vspace{-3mm}
\label{boxplot}
\end{figure*}

\begin{figure*}[t]
\includegraphics[width=\textwidth]{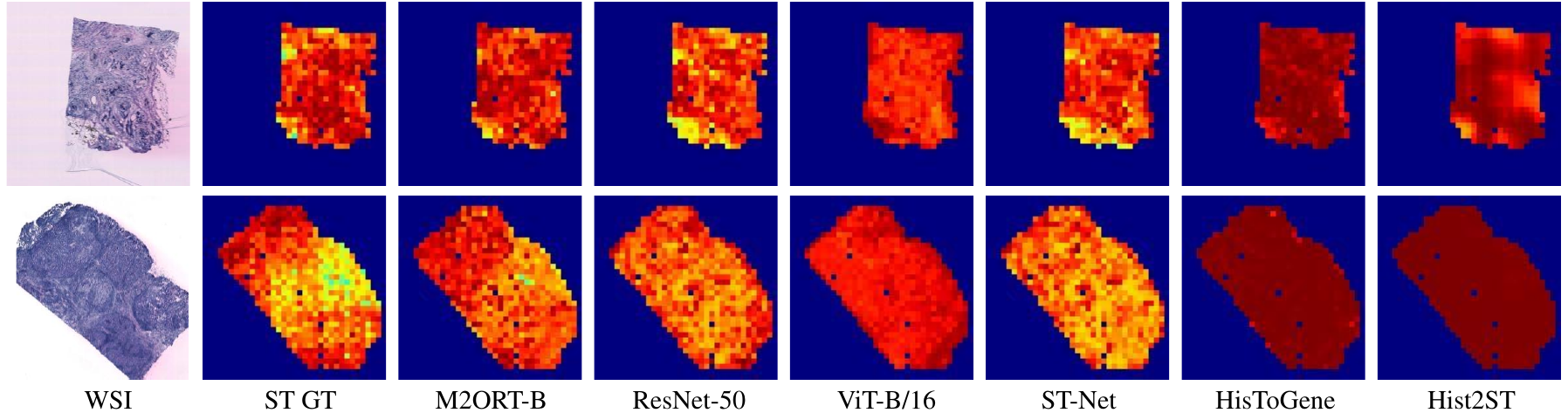}
\caption{Visualization results of the predicted ST maps on the test set of HBC datasets.} 
% \vspace{-3mm}
\label{visualization_st}
\end{figure*}

\subsection{Experimental Results}

\subsubsection{Comparison of M2ORT Series and Other Methods}
The experimental results of the comparison between many-to-one-based M2ORT series and other one-to-one-based methods are presented in Table~\ref{comparison_with_other_methods}. 
% This table provides detailed insights into the PCC and RMSE on various datasets of different methods, along with their parameter count and FLOPs. 
% It should be noted that the FLOPs of HisToGene and Hist2ST may vary depending on the dataset. This is because HisToGene and Hist2ST are slide-level methods, and their FLOPs are influenced by the length of sequences. Consequently, the table reports the highest GPU memory consumption among all three datasets for these two methods.
Analysis of the experimental results reveals that M2ORT series achieve superior performance with fewer FLOPs and a reduced parameter count. Across all three datasets, M2ORT-L attains the highest PCC and the lowest RMSE. Meanwhile, M2ORT-B secures the second-highest performance with fewer parameters and FLOPs. Additionally, M2ORT-S also exhibits impressive performance with the most lightweight design among the considered methods. In comparison to ST-Net, which features 0.38 million more parameters and 1.43G more FLOPs, M2ORT-S surpasses its performance on HER2+ and cSCC datasets by 0.97\% and 1.23\% PCC, respectively.

Additionally, a visual comparison between these methods is presented in Figure~\ref{performance_visual_comparison}, where their performance is evaluated based on the mean PCC across the three datasets. Notably, M2ORT-S outperforms all other methods in terms of mean PCC and achieves significant savings of GPU computational cost during training, making it the most cost-effective method among all the methods. These results underscore the effectiveness and efficiency of the proposed M2ORT series.

\subsubsection{Comparison of Patch-Level and Slide-Level Methods}

From Table~\ref{comparison_with_other_methods}, it is also observed that the slide-level methods fail to outperform patch-level methods on all three datasets. Among the slide-level methods, Hist2ST does surpass HisToGene due to its larger model size, but the extra FLOPs and the dramatic parameter count diminish the significance of this performance improvement. When compared to baseline patch-level methods such as ResNet-50, the PCC of Hist2ST is 2.10\%, 3.31\%, and 2.63\% lower on the three datasets respectively. This suggests that the gene expressions of a spot are primarily related to its corresponding tissue area, and introducing inter-patch correlations does little to enhance prediction accuracy.
Nevertheless, slide-level methods still possess the advantage of being more efficient in generating entire ST maps. With a refined network design, they still have the potential of achieving a competitive regression accuracy.

\subsubsection{Stability Analysis}

To assess the regression stability of various methods, we further conducted a comprehensive box-plot comparison, as depicted in Figure~\ref{boxplot}. Comparing the three primary patch-level regression baselines, namely ResNet-50, ViT-B/16, and ST-Net, it is shown that ST-Net showcased superior overall stability in comparison to the other methods despite exhibiting a higher number of outliers. Transitioning to slide-level methods HisToGene and Hist2ST, the stability experienced a notable decline, particularly evident in the cSCC dataset. Lastly, introducing M2ORT-B into the comparison revealed that the proposed M2ORT method demonstrated the most stable predictions across all considered methods, validating the effectiveness of our network design.

% \subsection{Study on the Transferring Ability}

% We also tested the transferring ability of different methods to make sure they can adapt to unseen data in real-life clinical use.

\subsubsection{Visualization Analysis}

Finally, we present some visualization results in Figure~\ref{visualization_st} to have an intuitive comparison of the methods. Principal Component Analysis is used here to compress the 250-dimension gene expressions into one dimension for better color mapping and visualization. As it is shown, slide-level methods such as HisToGene and Hist2ST tend to generate smoother ST maps, owing to the holistic processing of entire slides. In contrast, patch-level methods typically yield sharper predictions due to the independent processing of each spot in the ST map. Notably, M2ORT consistently produces more accurate ST maps with distributions closely resembling the ground truth. This observation underscores the effectiveness of the many-to-one design implemented in M2ORT.

\section{Conclusion}

In this study, we tackle the challenging task of predicting ST gene expressions by proposing a novel many-to-one-based regression Transformer, M2ORT. M2ORT leverages pathology images from three distinct levels to collectively predict gene expressions within their common corresponding tissue area. The model incorporates M2ORT Encoder for decoupled multi-scale feature extraction, which comprises ITMM for intra-scale representation learning and ICMM for cross-scale feature fusion. The experimental results on three public datasets show that M2ORT series can achieve state-of-the-art performance with fewer parameters and FLOPs.

\bibliographystyle{named}
\bibliography{ijcai24}

\begin{thebibliography}{}

\bibitem[\protect\citeauthoryear{Andersson \bgroup \em et al.\egroup }{2021}]{dataset2}
Alma Andersson, Ludvig Larsson, Linnea Stenbeck, Fredrik Salm{\'e}n, Anna Ehinger, Sunny~Z Wu, Ghamdan Al-Eryani, Daniel Roden, Alex Swarbrick, {\AA}ke Borg, et~al.
\newblock Spatial deconvolution of her2-positive breast cancer delineates tumor-associated cell type interactions.
\newblock {\em Nature communications}, 12(1):6012, 2021.

\bibitem[\protect\citeauthoryear{Ba \bgroup \em et al.\egroup }{2016}]{layernorm}
Jimmy~Lei Ba, Jamie~Ryan Kiros, and Geoffrey~E Hinton.
\newblock Layer normalization.
\newblock {\em arXiv preprint arXiv:1607.06450}, 2016.

\bibitem[\protect\citeauthoryear{Bressan \bgroup \em et al.\egroup }{2023}]{bressan2023dawn}
Dario Bressan, Giorgia Battistoni, and Gregory~J Hannon.
\newblock The dawn of spatial omics.
\newblock {\em Science}, 381(6657):eabq4964, 2023.

\bibitem[\protect\citeauthoryear{Cang \bgroup \em et al.\egroup }{2023}]{cang2023screening}
Zixuan Cang, Yanxiang Zhao, Axel~A Almet, Adam Stabell, Raul Ramos, Maksim~V Plikus, Scott~X Atwood, and Qing Nie.
\newblock Screening cell--cell communication in spatial transcriptomics via collective optimal transport.
\newblock {\em Nature Methods}, 20(2):218--228, 2023.

\bibitem[\protect\citeauthoryear{Chen \bgroup \em et al.\egroup }{2022}]{hipt}
Richard~J Chen, Chengkuan Chen, Yicong Li, Tiffany~Y Chen, Andrew~D Trister, Rahul~G Krishnan, and Faisal Mahmood.
\newblock Scaling vision transformers to gigapixel images via hierarchical self-supervised learning.
\newblock In {\em Proceedings of the IEEE/CVF Conference on Computer Vision and Pattern Recognition}, pages 16144--16155, 2022.

\bibitem[\protect\citeauthoryear{Ding \bgroup \em et al.\egroup }{2023}]{ding2023pathology}
Kexin Ding, Mu~Zhou, Dimitris~N Metaxas, and Shaoting Zhang.
\newblock Pathology-and-genomics multimodal transformer for survival outcome prediction.
\newblock In {\em International Conference on Medical Image Computing and Computer-Assisted Intervention}, pages 622--631. Springer, 2023.

\bibitem[\protect\citeauthoryear{Dosovitskiy \bgroup \em et al.\egroup }{2020}]{vit}
Alexey Dosovitskiy, Lucas Beyer, Alexander Kolesnikov, Dirk Weissenborn, Xiaohua Zhai, Thomas Unterthiner, Mostafa Dehghani, Matthias Minderer, Georg Heigold, Sylvain Gelly, et~al.
\newblock An image is worth 16x16 words: Transformers for image recognition at scale.
\newblock In {\em International Conference on Learning Representations}, 2020.

\bibitem[\protect\citeauthoryear{Hamilton \bgroup \em et al.\egroup }{2017}]{hamilton2017inductive}
Will Hamilton, Zhitao Ying, and Jure Leskovec.
\newblock Inductive representation learning on large graphs.
\newblock {\em Advances in neural information processing systems}, 30, 2017.

\bibitem[\protect\citeauthoryear{He \bgroup \em et al.\egroup }{2016}]{resnet}
Kaiming He, Xiangyu Zhang, Shaoqing Ren, and Jian Sun.
\newblock Deep residual learning for image recognition.
\newblock In {\em Proceedings of the IEEE Conference on Computer Vision and Pattern Recognition}, pages 770--778, 2016.

\bibitem[\protect\citeauthoryear{He \bgroup \em et al.\egroup }{2020}]{stnet}
Bryan He, Ludvig Bergenstr{\aa}hle, Linnea Stenbeck, Abubakar Abid, Alma Andersson, {\AA}ke Borg, Jonas Maaskola, Joakim Lundeberg, and James Zou.
\newblock Integrating spatial gene expression and breast tumour morphology via deep learning.
\newblock {\em Nature biomedical engineering}, 4(8):827--834, 2020.

\bibitem[\protect\citeauthoryear{Huang \bgroup \em et al.\egroup }{2017}]{densenet}
Gao Huang, Zhuang Liu, Laurens Van Der~Maaten, and Kilian~Q Weinberger.
\newblock Densely connected convolutional networks.
\newblock In {\em Proceedings of the IEEE conference on computer vision and pattern recognition}, pages 4700--4708, 2017.

\bibitem[\protect\citeauthoryear{Ji \bgroup \em et al.\egroup }{2020}]{dataset3}
Andrew~L Ji, Adam~J Rubin, Kim Thrane, Sizun Jiang, David~L Reynolds, Robin~M Meyers, Margaret~G Guo, Benson~M George, Annelie Mollbrink, Joseph Bergenstr{\aa}hle, et~al.
\newblock Multimodal analysis of composition and spatial architecture in human squamous cell carcinoma.
\newblock {\em Cell}, 182(2):497--514, 2020.

\bibitem[\protect\citeauthoryear{Kingma and Ba}{2015}]{adam}
Diederik~P Kingma and Jimmy Ba.
\newblock Adam: A method for stochastic optimization.
\newblock In {\em International Conference on Learning Representations}, 2015.

\bibitem[\protect\citeauthoryear{Kolodziejczyk \bgroup \em et al.\egroup }{2015}]{singlecellsequencing}
Aleksandra~A Kolodziejczyk, Jong~Kyoung Kim, Valentine Svensson, John~C Marioni, and Sarah~A Teichmann.
\newblock The technology and biology of single-cell rna sequencing.
\newblock {\em Molecular cell}, 58(4):610--620, 2015.

\bibitem[\protect\citeauthoryear{Lee \bgroup \em et al.\egroup }{2021}]{lee2021xyzeq}
Youjin Lee, Derek Bogdanoff, Yutong Wang, George~C Hartoularos, Jonathan~M Woo, Cody~T Mowery, Hunter~M Nisonoff, David~S Lee, Yang Sun, James Lee, et~al.
\newblock Xyzeq: Spatially resolved single-cell rna sequencing reveals expression heterogeneity in the tumor microenvironment.
\newblock {\em Science advances}, 7(17):eabg4755, 2021.

\bibitem[\protect\citeauthoryear{Levy-Jurgenson \bgroup \em et al.\egroup }{2020}]{levy2020spatial}
Alona Levy-Jurgenson, Xavier Tekpli, Vessela~N Kristensen, and Zohar Yakhini.
\newblock Spatial transcriptomics inferred from pathology whole-slide images links tumor heterogeneity to survival in breast and lung cancer.
\newblock {\em Scientific reports}, 10(1):18802, 2020.

\bibitem[\protect\citeauthoryear{Monjo \bgroup \em et al.\egroup }{2022}]{deepspace}
Taku Monjo, Masaru Koido, Satoi Nagasawa, Yutaka Suzuki, and Yoichiro Kamatani.
\newblock Efficient prediction of a spatial transcriptomics profile better characterizes breast cancer tissue sections without costly experimentation.
\newblock {\em Scientific Reports}, 12(1):4133, 2022.

\bibitem[\protect\citeauthoryear{Mrabah \bgroup \em et al.\egroup }{2023}]{mrabah2023toward}
Nairouz Mrabah, Mohamed~Mahmoud Amar, Mohamed Bouguessa, and Abdoulaye~Banire Diallo.
\newblock Toward convex manifolds: a geometric perspective for deep graph clustering of single-cell rna-seq data.
\newblock In {\em Proceedings of the Thirty-Second International Joint Conference on Artificial Intelligence}, pages 4855--4863, 2023.

\bibitem[\protect\citeauthoryear{Niazi \bgroup \em et al.\egroup }{2019}]{niazi2019digital}
Muhammad Khalid~Khan Niazi, Anil~V Parwani, and Metin~N Gurcan.
\newblock Digital pathology and artificial intelligence.
\newblock {\em The lancet oncology}, 20(5):e253--e261, 2019.

\bibitem[\protect\citeauthoryear{Pang \bgroup \em et al.\egroup }{2021a}]{histogene}
Minxing Pang, Kenong Su, and Mingyao Li.
\newblock Leveraging information in spatial transcriptomics to predict super-resolution gene expression from histology images in tumors.
\newblock {\em bioRxiv}, pages 2021--11, 2021.

\bibitem[\protect\citeauthoryear{Pang \bgroup \em et al.\egroup }{2021b}]{pang2021leveraging}
Minxing Pang, Kenong Su, and Mingyao Li.
\newblock Leveraging information in spatial transcriptomics to predict super-resolution gene expression from histology images in tumors.
\newblock {\em bioRxiv}, pages 2021--11, 2021.

\bibitem[\protect\citeauthoryear{Rao \bgroup \em et al.\egroup }{2021}]{rao2021exploring}
Anjali Rao, Dalia Barkley, Gustavo~S Fran{\c{c}}a, and Itai Yanai.
\newblock Exploring tissue architecture using spatial transcriptomics.
\newblock {\em Nature}, 596(7871):211--220, 2021.

\bibitem[\protect\citeauthoryear{Rodriques \bgroup \em et al.\egroup }{2019}]{rodriques2019slide}
Samuel~G Rodriques, Robert~R Stickels, Aleksandrina Goeva, Carly~A Martin, Evan Murray, Charles~R Vanderburg, Joshua Welch, Linlin~M Chen, Fei Chen, and Evan~Z Macosko.
\newblock Slide-seq: A scalable technology for measuring genome-wide expression at high spatial resolution.
\newblock {\em Science}, 363(6434):1463--1467, 2019.

\bibitem[\protect\citeauthoryear{Ryu \bgroup \em et al.\egroup }{2023}]{ryu2023ocelot}
Jeongun Ryu, Aaron~Valero Puche, JaeWoong Shin, Seonwook Park, Biagio Brattoli, Jinhee Lee, Wonkyung Jung, Soo~Ick Cho, Kyunghyun Paeng, Chan-Young Ock, et~al.
\newblock Ocelot: Overlapped cell on tissue dataset for histopathology.
\newblock In {\em Proceedings of the IEEE/CVF Conference on Computer Vision and Pattern Recognition}, pages 23902--23912, 2023.

\bibitem[\protect\citeauthoryear{Simonyan and Zisserman}{2015}]{vgg}
K~Simonyan and A~Zisserman.
\newblock Very deep convolutional networks for large-scale image recognition.
\newblock In {\em 3rd International Conference on Learning Representations (ICLR 2015)}. Computational and Biological Learning Society, 2015.

\bibitem[\protect\citeauthoryear{Song and Su}{2021}]{song2021dstg}
Qianqian Song and Jing Su.
\newblock Dstg: deconvoluting spatial transcriptomics data through graph-based artificial intelligence.
\newblock {\em Briefings in bioinformatics}, 22(5):bbaa414, 2021.

\bibitem[\protect\citeauthoryear{Srivastava \bgroup \em et al.\egroup }{2014}]{dropout}
Nitish Srivastava, Geoffrey Hinton, Alex Krizhevsky, Ilya Sutskever, and Ruslan Salakhutdinov.
\newblock Dropout: a simple way to prevent neural networks from overfitting.
\newblock {\em The journal of machine learning research}, 15(1):1929--1958, 2014.

\bibitem[\protect\citeauthoryear{Stenbeck \bgroup \em et al.\egroup }{2021}]{dataset1}
Linnea Stenbeck, L~Bergenstr{\aa}hle, J~Lundeberg, and {\AA}~Borg.
\newblock Human breast cancer in situ capturing transcriptomics.
\newblock {\em Mendeley Data}, 2, 2021.

\bibitem[\protect\citeauthoryear{Tian \bgroup \em et al.\egroup }{2020}]{tian2020brd2}
Xiao-Peng Tian, Jun Cai, Shu-Yun Ma, Yu~Fang, Hui-Qiang Huang, Tong-Yu Lin, Hui-Lan Rao, Mei Li, Zhong-Jun Xia, Tie-Bang Kang, et~al.
\newblock Brd2 induces drug resistance through activation of the rasgrp1/ras/erk signaling pathway in adult t-cell lymphoblastic lymphoma.
\newblock {\em Cancer Communications}, 40(6):245--259, 2020.

\bibitem[\protect\citeauthoryear{Tian \bgroup \em et al.\egroup }{2023}]{tian2023expanding}
Luyi Tian, Fei Chen, and Evan~Z Macosko.
\newblock The expanding vistas of spatial transcriptomics.
\newblock {\em Nature Biotechnology}, 41(6):773--782, 2023.

\bibitem[\protect\citeauthoryear{Vaswani \bgroup \em et al.\egroup }{2017}]{transformer}
Ashish Vaswani, Noam Shazeer, Niki Parmar, Jakob Uszkoreit, Llion Jones, Aidan~N Gomez, {\L}ukasz Kaiser, and Illia Polosukhin.
\newblock Attention is all you need.
\newblock {\em Advances in neural information processing systems}, 30, 2017.

\bibitem[\protect\citeauthoryear{Wang \bgroup \em et al.\egroup }{2015}]{wang2015ddx}
Zhendong Wang, Zhonghua Luo, Lin Zhou, Xiaofei Li, Tao Jiang, and Enqing Fu.
\newblock Ddx 5 promotes proliferation and tumorigenesis of non-small-cell lung cancer cells by activating $\beta$-catenin signaling pathway.
\newblock {\em Cancer science}, 106(10):1303--1312, 2015.

\bibitem[\protect\citeauthoryear{Weitz \bgroup \em et al.\egroup }{2021}]{weitz2021investigation}
Philippe Weitz, Yinxi Wang, Johan Hartman, and Mattias Rantalainen.
\newblock An investigation of attention mechanisms in histopathology whole-slide-image analysis for regression objectives.
\newblock In {\em Proceedings of the IEEE/CVF International Conference on Computer Vision}, pages 611--619, 2021.

\bibitem[\protect\citeauthoryear{Yarlagadda \bgroup \em et al.\egroup }{2023}]{yarlagadda2023discrete}
Dig Vijay~Kumar Yarlagadda, Joan Massagu{\'e}, and Christina Leslie.
\newblock Discrete representation learning for modeling imaging-based spatial transcriptomics data.
\newblock In {\em Proceedings of the IEEE/CVF International Conference on Computer Vision}, pages 3846--3855, 2023.

\bibitem[\protect\citeauthoryear{Zeng \bgroup \em et al.\egroup }{2022}]{hist2st}
Yuansong Zeng, Zhuoyi Wei, Weijiang Yu, Rui Yin, Yuchen Yuan, Bingling Li, Zhonghui Tang, Yutong Lu, and Yuedong Yang.
\newblock Spatial transcriptomics prediction from histology jointly through transformer and graph neural networks.
\newblock {\em Briefings in Bioinformatics}, 23(5):bbac297, 2022.

\end{thebibliography}

\newpage
\section{Appendix}
\appendix

\begin{figure}[t]
\includegraphics[width=0.48\textwidth]{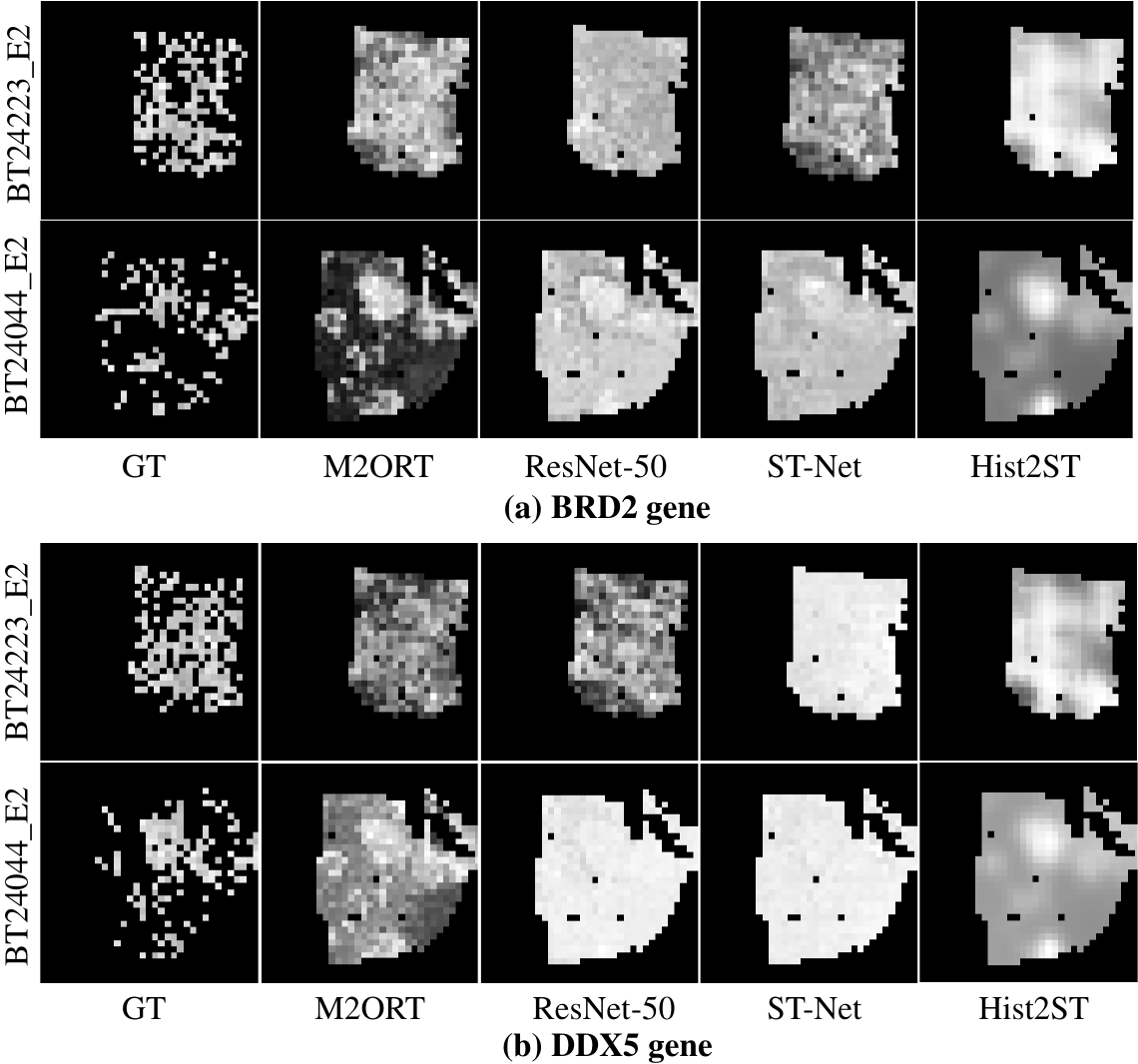}
\caption{(a) Visualization of the spatial distribution of the BRD2 gene. (b) Visualization of the spatial distribution of the DDX5 gene.} 
% \vspace{-3mm}
\label{per_gene_visualization}
\end{figure}

\section{Visualization of Some Individual Genes}

In addition to the Principal Component Analysis-based visualization results presented in the main paper, we augment our findings with supplementary individual gene visualizations in the Supplementary Materials to further elucidate the efficacy of M2ORT. The individual gene visualizations are generated with the Image Processing Toolbox in Matlab.

As depicted in Fig~\ref{per_gene_visualization}, we showcase the individual gene visualization maps of ResNet-50 \cite{resnet}, ST-Net \cite{stnet}, Hist2ST \cite{hist2st}, and M2ORT. We selected these methods for comparison because ResNet-50 and ST-Net demonstrated notable performance in our evaluations in the main paper, while Hist2ST exhibited the highest Pearson Correlation Coefficient (PCC) among slide-level methods. Two Whole Slide Images (WSIs) are used for visualization, namely BT24044\_E2 and BT24223\_E2. The genes we selected for visualization are BRD2 and DDX5. The genes chosen for visualization are BRD2 and DDX5. Notably, BRD2 induces drug resistance through the activation of the RasGRP1/Ras/ERK signaling pathway in adult T-cell lymphoblastic lymphoma \cite{tian2020brd2}, while DDX5 plays a pivotal role in the proliferation and tumorigenesis of non-small-cell cancer cells by activating the beta-catenin signaling pathway \cite{wang2015ddx}.

Our results indicate that M2ORT achieves the highest accuracy in gene expression prediction for the selected genes in the two WSIs, surpassing the performance of other patch-level and slide-level methods. Consistent with the visualization results presented in the main paper, slide-level methods continue to exhibit a tendency to generate smoother Spatial Transcriptomics (ST) maps for these individual genes.

\section{Details about the Data Structure}

\subsection{Details about the Data Structure of WSIs}

WSI files are organized in a hierarchical pyramidal structure, with each pyramid level representing a pathology image scanned from the glass slide at a distinct magnification factor. Pathology images at level 0 typically constitute gigapixel images, providing intricate cell-level details, while higher levels encapsulate more region-level morphological information \cite{hipt}. In routine clinical practice, pathologists typically commence their analysis from level 0 and progressively transition to images with reduced magnifications for a broader assessment of bio-markers \cite{ryu2023ocelot}. In the context of M2ORT, we assume that level 1 images are two times smaller than their level 0 counterparts. However, it's noteworthy that some WSIs generated by other systems may exhibit a four times reduction in size for level 1 images. The ablation study results presented in our paper demonstrate a noteworthy observation – the model's performance improvement persists even when utilizing the level 0 image and the four times smaller level 1 image as inputs. Importantly, we find that employing bilinear interpolation to synthesize a two-times down-sampled image, in conjunction with the original level 0 and reduced level 1 images, can yield comparable results. Our experiments reveal that this approach maintains a similar regression accuracy, notwithstanding the subtle distinctions between the synthesized and authentic images.

\subsection{Details about the File Structure of ST Data}

ST technologies generate spatially-aware gene expression maps based on the mRNA transcripts information. In addition to the corresponding WSI, each case's ST data also encompasses a minimum of two matrices designed for pathological analysis. One is the gene expression matrix, of which each column is a unique gene and each row represents an ST spot (or vice versa) indexed by its position in the ST map. The second matrix serves the purpose of spot-pixel mapping, featuring each row as a spot with its index, and columns detailing the pixel position of this spot in the level 0 pathology image. Optionally, there also could be a spot-level annotation matrix, where the category of each spot is recorded. However, since our work focuses on the prediction of the ST maps, such spot-level annotation matrices are not used in our experiments.

\section{Details about the M2ORT Model}

This study primarily centers on introducing a novel perspective to model the ST prediction problem. Therefore, our contributions are more about offering a comprehensive solution to this problem, with a focus on the overall methodology. We did not introduce many intricate network designs in M2ORT, and the methodology section in the main paper is written to be concise for better understanding. However, for a more in-depth understanding of the M2ORT architecture, we provide detailed insights into its structural elements in this section.

\subsection{Details about LDPE}
LDPE uses different patch sizes during patch embedding for different levels of pathology images. Besides that, each level's LDPE also differs in the network width for more efficient patch embedding. For example, LDPE of level 0 aims at project the $p\times p\times 3$ patches into $1\times C$ embedding tokens while LDPE of level 2 needs to project the $\frac{p}{4}\times \frac{p}{4}\times 3$ patches into $1\times C$ embedding tokens, where $p$ is the patch size for patch embedding on level 0, $C$ is the embedding dimension. In this case, LDPE of level 0 can be defined as:
\begin{equation}
    P_0=Rearrange(I_0) \in R^{\frac{H}{p}\times \frac{W}{p} \times p \times p \times 3},
\end{equation}
\begin{equation}
    h_0=GELU(Linear(Norm(P_0))) \in R^{\frac{H}{p}\times \frac{W}{p} \times 3C},
\end{equation}
\begin{equation}
    S_0=Norm(Linear(h_0)) \in R^{\frac{H}{p}\times \frac{W}{p} \times C},
\end{equation}
\noindent where $I_0$ is the level 0 pathology image, $h_0$ is the latent space representation, and $S_0$ is the output sequence for level 0. 

LDPE of level 1 and level 2 also have a similar process as aforementioned. But given that $P_1 \in R^{\frac{H/2}{p/2}\times \frac{W/2}{p/2} \times \frac{p}{2} \times \frac{p}{2} \times 3}$ and $P_2 \in R^{\frac{H/4}{p/4}\times \frac{W/4}{p/4} \times \frac{p}{4} \times \frac{p}{4} \times 3}$ have $4\times$ and $16\times$ fewer input elements for patch embedding, we adjust their latent space representation into $h_1 \in R^{\frac{H/2}{p/2}\times \frac{W/2}{p/2} \times \frac{3C}{2}}$ and $h_2 \in R^{\frac{H/4}{p/4}\times \frac{W/4}{p/4} \times \frac{3C}{4}}$, respectively. In the end, it is still ensured that $S_0$, $S_1$, and $S_2$ share the same shape of $L\times C$ through the second linear layer where $L=\frac{HW}{p^2}$.

\subsection{Details about M2ORT Encoder}

After LDPE, $[cls]$ tokens and positional encodings are added to the three sequences $S_0$, $S_1$, and $S_2$, respectively. Then, they are sent into M2ORT Encoder for multi-scale feature extraction, which is composed of ITMM for intra-scale feature extraction and ICMM for inter-scale feature fusion. Let us assume $n\_head$ as the number of heads used in ITMM, for an input sequence $S$, ITMM can be mathematically defined as:

\begin{equation}
    Q, K, V=Split(Linear(S)),
\end{equation}
\begin{equation}
    Q_i=Split(Q), \quad 1\le i\le n\_head
\end{equation}
\begin{equation}
    K_i=Split(K), \quad 1\le i\le n\_head
\end{equation}
\begin{equation}
    V_i=Split(V), \quad 1\le i\le n\_head
\end{equation}
\begin{equation}
    S_i=RMSA(Q_i,K_i,V_i,m),\ 1\le i\le n\_head
\end{equation}
\begin{equation}
    S_c=Concat({S_1, ..., S_i, ..., S_{n\_head}}),
\end{equation}
\begin{equation}
    S=Linear(S_c)+S,
\end{equation}

\noindent where $RMSA(\cdot)$ is the Random Mask Self-Attention that can be described as:

\begin{equation}
    RMSA(Q,K,V,m)=M(m)\odot Softmax(\frac{QK^T}{\sqrt{d_K}})V,
\end{equation}
\begin{equation}
\begin{aligned}
    & M(m)= &
    \begin{bmatrix}
    M^{(0),(0)} & ... & M^{(0),(j)} & ... & M^{(0),(L)}\\
    ... & ... & ... & ... & ... \\
    M^{(i),(0)} & ... & M^{(i),(j)} & ... & M^{(i),(L)}\\
    ... & ... & ... & ... & ... \\
    M^{(L),(0)} & ... & M^{(L),(j)} & ... & M^{(L),(L)}
    \end{bmatrix} \\
    & & s.t.\quad \left\{ 
    \begin{aligned}
    &P(M^{(i),(j)}=0)=m \cr
    &P(M^{(i),(j)}=1)=1-m
    \end{aligned}
    \right.
\end{aligned}
\end{equation}
% \begin{equation}
%     M^{(i),(j)}=\left\{ 
%     \begin{aligned}
%     &0 & &0\leqslant x<1 \cr 
%     &1 & &1\leqslant x<2
%     \end{aligned}
%     \right., 1\le i\le L+1, 1\le j\le L+1
% \end{equation}

Subsequently, ICMM can be defined as:

\begin{equation}
    S=Concat(S_0,S_1,S_2)\quad\in R^{(L+1)\times 3C},
\end{equation}
\begin{equation}
    S_c=GELU(Linear(S))\quad\in R^{(L+1)\times 2C},
\end{equation}
\begin{equation}
    S=Linear(S_c)+S\quad\in R^{(L+1)\times 3C},
\end{equation}
\begin{equation}
    S_0,S_1,S_2=Split(S).
\end{equation}

Then, M2ORT Encoder can be defined as:

\begin{equation}
    S_0, S_1, S_2=Split(Norm(Concat(S_0,S_1,S_2))),
\end{equation}
\begin{equation}
    S_i=ITMM(S_i),\quad 0\le i\le 2
\end{equation}
\begin{equation}
    S_0, S_1, S_2=Split(Norm(Concat(S_0,S_1,S_2))),
\end{equation}
\begin{equation}
    S_0, S_1, S_2=ICMM(S_0,S_1,S_2))),
\end{equation}

\subsection{The Pseudo Code of M2ORT}
Finally, We provide the pseudo-code of M2ORT pipeline in Algorithm~\ref{alg:algorithm}, where $PE_i$ denotes the fully-learnable positional encoding for the $i$-th level input sequence.

\begin{algorithm}[tb]
    \caption{M2ORT for ST prediction}
    \label{alg:algorithm}
    \textbf{Input}: Pathology images of three levels $I_0$, $I_1$, $I_2$\\
    \textbf{Parameter}: the repeated times $N$ of M2ORT Encoder, number of head $n\_head$ in ITMM, and the channels $C$ for LDPE.\\
    \textbf{Output}: The gene expressions in the corresponding spot of the images
    \begin{algorithmic}[1] %[1] enables line numbers
        \STATE $S_0=Concat(LDPE_0(I_0|C),[cls])$.
        \STATE $S_1=Concat(LDPE_1(I_1|C),[cls])$.
        \STATE $S_2=Concat(LDPE_2(I_2|C),[cls])$.
        \STATE $S_0=S_0+PE_0$.
        \STATE $S_1=S_1+PE_1$.
        \STATE $S_2=S_2+PE_2$.
        \FOR{$i$ in range($N$)}
        \STATE $S_0, S_1, S_2=Split(Norm(Concat(S_0, S_1, S_2)))$
        \STATE $S_0=ITMM_0^i(S_0|n\_head)$.
        \STATE $S_1=ITMM_1^i(S_1|n\_head)$.
        \STATE $S_2=ITMM_2^i(S_2|n\_head)$.
        \STATE $S_0, S_1, S_2=Split(Norm(Concat(S_0, S_1, S_2)))$
        \STATE $S_0, S_1, S_2=ICMM^i(S_0, S_1, S_2)$
        \ENDFOR
        \STATE $S_0, S_1, S_2=Split(Norm(Concat(S_0, S_1, S_2)))$
        \STATE $[cls]=Concat(S_0[0,:],S_1[0,:],S_2[0,:])$
        \STATE $\hat{G}=Linear([cls])$
        \STATE \textbf{return} $\hat{G}$
    \end{algorithmic}
\end{algorithm}

\section{More about the Implementation Details}

\subsection{Details about the Gene Selection Method}

Due to the inherently sparse nature of the ST map, we need to filter out less-variable genes in each dataset before initiating the training process. Following previous works \cite{stnet}, we select 250 spatially variable genes for the regression task. These genes are selected by the following steps:

\begin{enumerate}
    \item Calculate the respective variance of each gene in a WSI, and select the genes with top-1000 variance in this WSI.
    \item Find the union set of these 1000 genes, which will usually lead to a little more than 250 final candidates.
    \item Randomly sample 250 genes from the union set, and consistently utilize this subset for all subsequent experiments.
\end{enumerate}

In practical applications, doctors have the flexibility to manually specify the target genes before the training process. This enables predictions tailored to their specific analysis needs, enhancing the utility of the model in real-life scenarios.

\subsection{Details about the Evaluation of GPU Memory Consumption}

The training GPU memory is computed based on a batch size of 96 for patch-level methods and 1 for slide-level methods. However, it's noteworthy that the two slide-level methods, namely HisToGene and Hist2ST, exhibit variable GPU memory consumption during training. This variability is attributed to the fact that slide-level methods utilize the entire ST spots as sequential input. Consequently, as the number of available spots changes, the GPU memory consumption during training also fluctuates accordingly. To ensure a fair comparison, we record the highest GPU memory consumption observed during training across the three datasets for these slide-level methods.

\begin{figure*}[t]
\includegraphics[width=\textwidth]{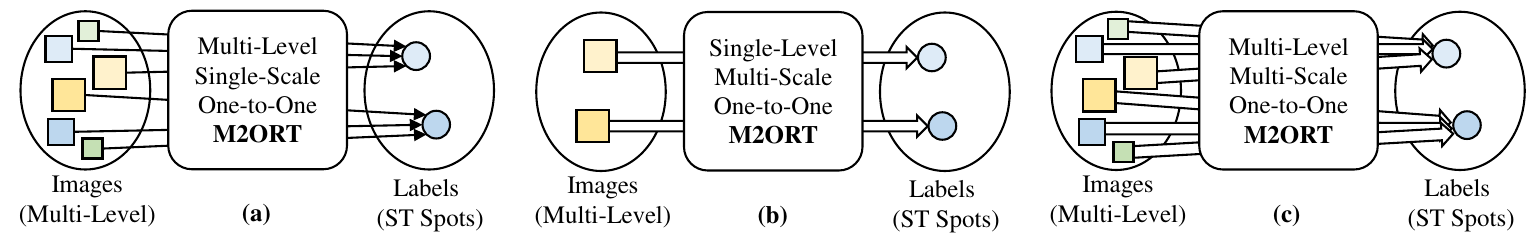}
\caption{The pipeline of some other one-to-one-based methods. All these methods only accept one single image as input at a time, and have an inferior performance compared to our proposed many-to-one-based methods. (a) The pipeline of multi-level single-scale one-to-one methods. (b) The pipeline of single-level multi-scale one-to-one methods. (c) The pipeline of multi-level multi-scale one-to-one methods. } 
% \vspace{-3mm}
\label{m2o_vs_o2o_multi-scale}
\end{figure*}

\section{More Experimental Results}

\begin{table}[t]
\renewcommand\tabcolsep{1pt}
\center
\begin{tabular}{@{}ccccccccc@{}}
\toprule
\multirow{2}{*}{Variant}  & \multirow{2}{*}{ITMM} & \multirow{2}{*}{ICMM} &\multicolumn{2}{c}{HBC} &\multicolumn{2}{c}{HER2+} &\multicolumn{2}{c}{cSCC} \\ \cmidrule(lr){4-5} \cmidrule(lr){6-7}\cmidrule(lr){8-9}
& &  & PCC   & RMSE  &   PCC   &   RMSE   &   PCC   &   RMSE    \\ \midrule
\multirow{3}{*}{Small} & \checkmark  & \checkmark & 47.78 &3.13 &43.98& 2.89 &50.70& 3.38   \\
 & \checkmark  & $\times$ & 45.62    &   3.21        &  41.89       & 3.19  &  48.19  &  3.75               \\ 
  & $\times$  & \checkmark & 43.77    &   3.29       &  40.54      &  3.22   & 47.44   &  3.79                 \\ \midrule 
\multirow{3}{*}{Base} & \checkmark  & \checkmark & 48.59  &  3.13       & 44.37       & 2.89 & 50.97  &3.37   \\
 & \checkmark  & $\times$ & 46.39    &  3.18       &  42.21   & 3.15   &  48.77 &  3.73   \\ 
  & $\times$  & \checkmark & 44.42    &   3.24       &  41.19      &  3.18   &  47.99  & 3.76                  \\ \midrule 
 \multirow{3}{*}{Large} & \checkmark  & \checkmark & 48.90 &3.10 &44.61& 2.87 &51.21 &3.34   \\
 & \checkmark  & $\times$ & 47.02   & 3.15  &  45.13  & 3.10       & 50.11    &  3.69                        \\ 
  & $\times$  & \checkmark & 46.88  & 3.16  &   44.89       &  3.12          & 49.44   &  3.72                     \\\bottomrule
\end{tabular}
\caption{Ablation study based on the removal of components. PCC is reported in percentage numbers.}
\label{ablation_study}
\end{table}

\subsection{Additional Ablation Study on M2ORT Encoder}
\label{sec_exp_o2o}
M2ORT boasts a highly compact model design, and the performance of the model is significantly influenced when certain components are directly removed. Therefore, the ablation studies presented in the main paper primarily concentrate on validating the effectiveness of the decoupling mechanism within the M2ORT Encoder, specifically based on its two components, ITMM and ICMM. Within the main paper, the ablation study is executed by transforming these components into non-decoupled implementations, underscoring the superiority of our model design. In this supplementary material, we provide additional ablation study results where each module is directly discarded, aiming to verify its effectiveness. The detailed results are presented in Table~\ref{ablation_study}.

We initiate the ablation study by removing ICMM from M2ORT. In this scenario, the M2ORT Encoder lacks cross-scale information exchange, and the three $[cls]$ tokens are directly concatenated for ST spot prediction. Our findings indicate a decline in model performance across all datasets. Notably, the performance in this case even falls short of the scenario where only level 0 images are used as input (as presented in Table 2 in the main paper). This is attributed to the identical network width and depth for both cases. When utilizing images from all three levels as input, the learning space for level 0 input is reduced to one-third of the scheme using level 0 images only.

Subsequently, we reinstate ICMM and proceed to remove ITMM from M2ORT. This time, the performance experiences a more substantial drop, as the intra-scale features cannot be fully exploited. In this case, all feature extraction and cross-scale information exchange operations rely solely on the two linear layers in ICMM, resulting in an under-fitting problem. In summary, the ablation study results underscore the indispensability of both ITMM and ICMM as integral components for effective multi-scale feature extraction in M2ORT.

\subsection{More Experiments on One-To-One Methods}

In this section, we delve into a more detailed ablation study regarding the many-to-one scheme proposed in M2ORT, exploring comparisons from three distinct perspectives: training sample pool (samples from multiple levels or a single level), feature extraction scheme (employment of multi-scale feature extraction methods or not), and pipeline design (many-to-one or one-to-one).

In scenarios where multi-level inputs are employed for training, the model can harness multi-scale information through many-to-one-based methods or attempt to extract multi-scale information within each individual image using special designs in one-to-one-based methods. Conversely, when single-level inputs are used for training, only one-to-one methods can be applied for modeling, as there are no multi-level samples for training many-to-one models. However, in this case, multi-scale information may still be extracted through specialized network designs. In the main paper, we have already presented the comparison results between M2ORT and single-level single-scale one-to-one methods in section~\ref{sec:m2o_v_o2v}. Here, based on M2ORT-B, we introduce additional comparisons involving single-level/multi-level single-scale/multi-scale one-to-one schemes for a more comprehensive analysis. The experimental results are detailed in Table~\ref{experimental_results}. And it is worth noting that, as all the compared methods in this section are one-to-one methods, the model only accepts a single image at a time even when multi-level images (i.e., images from multiple levels) are used for training.

\subsubsection{Multi-Level Single-Scale One-To-One M2ORT}

In section~\ref{sec:m2o_v_o2v} of the main paper, the experimental results affirm that the many-to-one scheme employed in M2ORT surpasses the one-to-one scheme in ST prediction. To further validate its effectiveness, we conducted a comparison between many-to-one M2ORT and one-to-one M2ORT trained with a pool containing pathology images from multiple levels.

More specifically, the one-to-one M2ORT trained with pathology images from multiple levels uses the pipeline presented in Figure~\ref{m2o_vs_o2o_multi-scale}(a). It uses a combination of level 0, level 1, and level 2 pathology images as individual training samples, and they are randomly fed into the model during training. However, due to the inability of one-to-one M2ORT to adapt different patch sizes for each level's patch embedding process, a fixed patch size of 16 is set for the patch embedding of all levels' inputs. To ensure samples in the same batch possess the same resolution (as required by PyTorch), level 1 and level 2 images are upsampled to the resolution of level 0 images.

The performance of this training scheme is outlined in the second row of Table~\ref{experimental_results}. It achieves PCC values of 47.21\%, 43.59\%, and 49.91\% on the three datasets, respectively. This performance slightly outperforms the single-level single-scale one-to-one M2ORT, presented in the sixth row of the table. This observation underscores that utilizing different levels of images for training can enhance the model's generalization ability and, consequently, improve testing accuracy. However, since the multi-scale features are not used directly during back-propagation, the performance still falls short of the dedicated multi-scale methods.

\begin{table}[t]
\renewcommand\tabcolsep{1pt}
\center
\begin{tabular}{@{}ccccccccc@{}}
\toprule
\multirow{2}{*}{\begin{tabular}[c]{@{}c@{}}Multi-\\Level\end{tabular}} & \multirow{2}{*}{\begin{tabular}[c]{@{}c@{}}Multi-\\Scale\end{tabular}} & \multirow{2}{*}{\begin{tabular}[c]{@{}c@{}}Many-\\to-One\end{tabular}} &\multicolumn{2}{c}{HBC} &\multicolumn{2}{c}{HER2+} &\multicolumn{2}{c}{cSCC} \\ \cmidrule(lr){4-5} \cmidrule(lr){6-7}\cmidrule(lr){8-9}
                         &         &                   & PCC      & RMSE        &   PCC               &   RMSE       &   PCC               &   RMSE                       \\ \midrule
\checkmark & \checkmark  & \checkmark & 48.59  &  3.13  & 44.37   & 2.89 & 50.97  &   3.37                    \\
\checkmark & $\times$  & $\times$  & 47.21  &  3.15  &  43.59     & 3.04  & 49.91   & 3.51     \\
$\times$  & \checkmark & $\times$  & 47.91   & 3.14 &  43.88   & 3.03    &  50.21           &    3.43                 \\
\checkmark & \checkmark  & $\times$ &  47.97 & 3.14  & 43.91  & 3.03   &   50.19   & 3.43            \\ 
% \checkmark   & $\times$ & \checkmark & 46.39  &  3.18   &  42.21   & 3.15   &  48.77 &  3.73                   \\ 
\midrule
$\times$  & $\times$  & $\times$  & 47.10 & 3.15& 43.33 & 3.05& 49.34 & 3.60    \\
  \bottomrule
\end{tabular}
\caption{Extra ablation studies on one-to-one methods. }
\label{experimental_results}
\end{table}

\subsubsection{Single-Level Multi-Scale One-To-One M2ORT}

Besides the single-level single-scale one-to-one M2ORT described in section~\ref{sec:m2o_v_o2v} of the main paper, we have also implemented an additional variant known as single-level multi-scale one-to-one M2ORT. This variant follows the pipeline illustrated in Figure~\ref{m2o_vs_o2o_multi-scale}(b). The objective of this pipeline is to directly extract multi-scale features from the single-level level 0 inputs. This method maintains a structure similar to the original M2ORT but replaces LDPE with a multi-scale patch embedder designed to handle single-level inputs. The multi-scale patch embedder utilizes different patch sizes for patch embedding on the single-level inputs to extract multi-scale information. To facilitate cross-scale information exchange, resizing is employed to maintain sequences of different scales at the same length.

As is shown in the third row of the table, the incorporation of multi-scale design into the model results in a substantial improvement in performance across all three datasets. This robust enhancement underscores the effectiveness of utilizing multi-scale features for gene expression prediction.

\subsubsection{Multi-Level Multi-Scale One-To-One M2ORT}

Next, we transform the aforementioned multi-level single-scale one-to-one M2ORT into multi-level multi-scale one-to-one M2ORT by introducing the multi-scale feature extraction design used in single-level multi-scale one-to-one M2ORT. The pipeline of this scheme is depicted in Figure~\ref{experimental_results}(c). Similar to the multi-level single-scale one-to-one M2ORT, pre-processing necessitates resizing images from different levels to match the resolution of level 0 images. LDPE also employs a fixed 16-patch size for patch embedding due to the simplicity of the one-to-one-based network structure.

The performance of this method is recorded in the fourth line of the table. As shown, this method also achieves a competitive accuracy. But when compared to the single-level single-scale one-to-one M2ORT, it turns out that the introduction of multi-level inputs does not increase the model performance much. This is because the resizing operation in the single-level multi-scale scheme can indirectly generate similar representations as that of the level 1 and level 2 images, making the difference between multi-level-based and single-level-based multi-scale methods less significant.

\subsubsection{Single-Level Single-Scale One-To-One M2ORT}

The single-level single-scale one-to-one M2ORT has been introduced in section~\ref{sec:m2o_v_o2v} of the main paper. In this context, we present its performance in the fifth row of the table for a more convenient comparison.

\end{document}